\documentclass[sigconf]{acmart}
\usepackage{multirow}
\usepackage{url}
\usepackage{makecell}
\usepackage{colortbl}
\usepackage{xcolor}
\AtBeginDocument{%
  }

\copyrightyear{2026}
\acmYear{2026}
\setcopyright{cc}
\setcctype{by}
\acmConference[MM '26]{Proceedings of the 34th ACM International Conference on Multimedia}{November 10--14, 2026}{Rio de Janeiro, Brazil}
\acmBooktitle{Proceedings of the 34th ACM International Conference on Multimedia (MM '26), November 10--14, 2026, Rio de Janeiro, Brazil}
\acmDOI{10.1145/3767308.3835666}
\acmISBN{979-8-4007-2213-4/2026/11}

\begin{document}

\title{MARE: Multimodal Alignment and Reinforcement for Explainable Deepfake Detection via Vision-Language Models}

\author{Wenbo Xu}
\email{xuwb25@mail2.sysu.edu.cn}
\orcid{0009-0007-8702-4521}
\affiliation{
  \institution{School of Computer Science and Engineering, Sun Yat-sen University}
  \city{Guangzhou}
  \state{Guangdong}
  \country{China}
}

\author{Wei Lu}
\correspondingauthor
\orcid{0000-0002-4068-1766}
\email{luwei3@mail.sysu.edu.cn}
\affiliation{
  \institution{School of Computer Science and Engineering, Sun Yat-sen University}
  \city{Guangzhou}
  \state{Guangdong}
  \country{China}
}

\author{Xiangyang Luo}
\orcid{0000-0003-3225-4649}
\email{luoxy\_ieu@sina.com}
\affiliation{
  \institution{State Key Laboratory of Mathematical Engineering and Advanced Computing}
  \city{Zhengzhou}
  \state{Henan}
  \country{China}
} 

\renewcommand{\shortauthors}{Wenbo Xu, Wei Lu, and Xiangyang Luo}

\begin{abstract}
Deepfake detection is a widely researched topic that is crucial for combating the spread of malicious content, with existing methods mainly modeling the problem as a classification task or a localization task.
The rapid advancements in generative models impose new demands on Deepfake detection.
In this paper, we propose multimodal alignment and reinforcement for explainable Deepfake detection via vision-language models, termed MARE, which aims to enhance the accuracy and reliability of large vision-language models (VLMs) in Deepfake detection and reasoning.
Specifically, MARE designs comprehensive reward functions, incorporating reinforcement learning from human feedback (RLHF), to incentivize the generation of text-spatially aligned reasoning content that adheres to human preferences.
Besides, MARE introduces a forgery disentanglement module to capture intrinsic forgery traces from high-level facial semantics, thereby improving its authenticity detection capability.
We conduct thorough evaluations on the reasoning content generated by MARE.
Both quantitative and qualitative experimental results demonstrate that MARE achieves state-of-the-art performance in terms of accuracy and reliability.
\end{abstract}

\ccsdesc[500]{Computing methodologies}
\ccsdesc[500]{Computing methodologies~Artificial intelligence}
\ccsdesc[500]{Computing methodologies~Computer vision}
\ccsdesc[500]{Computing methodologies~Computer vision problems}

\keywords{Deepfake detection, large vision-language models, reinforcement learning, multimodal alignment}


\maketitle

\section{Introduction}
Concerns about the security of AI have accompanied its evolution from the beginning.
A recent study shows that more than half of new articles on the internet are AI-generated \cite{aiGeneratemore}, surpassing human-created content.
The proliferation of AI-generated content has varying degrees of impact on individuals, society, and nations.
To mitigate the misuse of generative AI on face data, Deepfake detection has emerged as a popular research topic in recent years.
Its primary purpose is to identify forged face images or videos, involving tasks such as binary authenticity detection \cite{yin2024fine}, spatial forgery localization \cite{huang2025sida}, and temporal forgery localization \cite{xu2025multimodal}.
\begin{figure}[t]
  \centering
  \includegraphics[width=\linewidth]{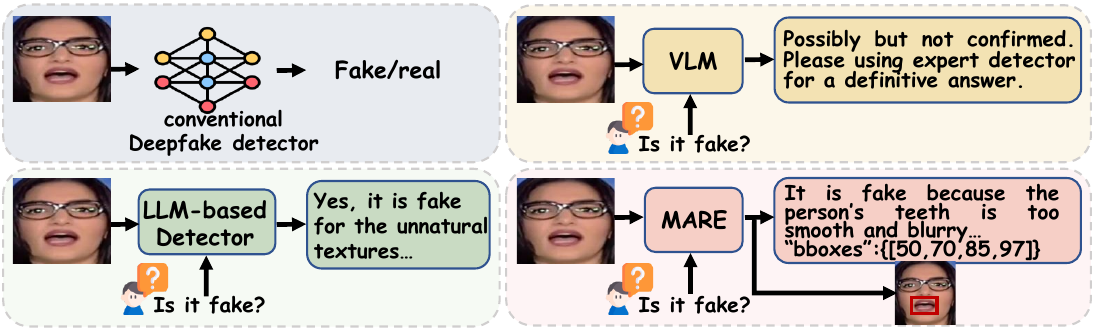}
  \caption{
      Illustration of different Deepfake detection methods. 
      The conventional Deepfake detectors provide a discriminative decision. 
      The existing LLM-based methods generate plain textual reasoning content. 
      While the pre-trained VLMs struggle to satisfy Deepfake reasoning demands.
      Our method generates detailed forgery traces analysis and provides spatial localization information for supporting evidence.
        }
  \label{fig_introduction}
\end{figure}

Existing Deepfake detection methods primarily exploit specific forgery traces within forged face data, such as blurred edges \cite{shiohara2022detecting}, spatial inconsistencies \cite{10702428}, and temporal jitter \cite{xu2025weakly}, to obtain a discriminative detection result.
With the rapid advancement of generative models, particularly large language models (LLMs) and large vision-language models (VLMs), new identification demands have emerged for Deepfake detection and reasoning.
Several recent methods have attempted to leverage LLMs for explainable Deepfake detection.
For example, M{2}F{2}-Det \cite{guo2025rethinking} enhances the explainability of Vicuna-7B \cite{chiang2023vicuna} by integrating the forgery prompt learning and the CLIP's multimodal representation ability.
KFD \cite{yu2025unlocking} proposes aligning face images with textual descriptions to generate forgery consistency maps, which achieves multimodal alignment through prompt learning.
RAIDX \cite{li2025raidx} integrates retrieval-augmented generation and group relative policy optimization to enhance the detection accuracy and decision explainability of LLMs.
The aforementioned methods primarily rely on combining prompt engineering with LLMs, which enhance the ability of explainable Deepfake detection through elaborate prompt design.
In this paper, we focus on directly leveraging VLMs for Deepfake detection.
By integrating reinforcement learning and multimodal alignment of reasoning content, we aim to enhance the accuracy and reliability of VLMs in Deepfake detection and reasoning.
VLMs exhibit superior performance in associating textual and visual modalities, thus generating more reliable content by maintaining multimodal representation alignment during the reasoning process \cite{zhai2024fine}.
While existing pre-trained VLMs perform well in scenarios involving strong semantic signals, such as visual question answering and image captioning, they tend to underperform in the task of Deepfake detection scenario \cite{yu2025unlocking, guo2025rethinking}.
The primary reason is that forged face images are often visually difficult to distinguish from genuine ones, since the forgery traces are typically subtle or imperceptible \cite{wang2025forensics}.
Therefore, it is necessary to enhance existing pre-trained VLMs for the specific scenario of Deepfake detection.

Reinforcement learning (RL) has been demonstrated to enhance the reasoning capability of VLMs through feedback on their outputs \cite{shen2025vlm}.
Compared to Supervised Fine-Tuning (SFT), the RL-based optimization strategy could mitigate model hallucinations and encourage the model to explore solutions that reflect human preferences.
By leveraging the reward mechanism in RL, we could not only incentivize the VLMs to capture subtle visual traces in forged face images, but also facilitate multimodal alignment in the reasoning process \cite{li2025survey}.
This significantly motivates the VLMs to explore and generate more accurate and reliable Deepfake detection and reasoning content.

In this paper, we propose MARE: multimodal alignment and reinforcement for explainable Deepfake detection via vision-language models.
The primary objective of MARE is to enhance the accuracy and reliability of VLMs in Deepfake detection and reasoning.
Under the reinforcement learning from human feedback (RLHF) paradigm, we first design a comprehensive reward mechanism.
Specifically, MARE designs multi-dimensional reward functions to incentivize the reasoning capability of VLMs in Deepfake detection, which include format, accuracy, text relevance, ROI, and alignment rewards (see Section \ref{sec:reward_func} for details).
To fulfill the data requirements for fine-tuning, a Deepfake multimodal alignment dataset is introduced by augmenting the existing image-text Deepfake dataset with aligned spatial localization information.
MARE also introduces a forgery disentanglement module to capture intrinsic forgery traces for precise authenticity identification, which explicitly disentangles the face image into identity, structural, and forgery traces features based on representation learning.
The main contributions are summarized as follows:
\begin{itemize}
  \item MARE employs a multimodal alignment strategy to enhance the reasoning capability of VLMs for Deepfake detection and reasoning, which incentivizes the VLMs to explore and generate more accurate and reliable reasoning content under RLHF paradigm.
  \item MARE introduces a novel forgery disentanglement module to capture intrinsic forgery traces within face images, which improves the authenticity detection capability through representation learning.
  \item MARE achieves state-of-the-art performance in Deepfake detection and reasoning, which has been validated by extensive quantitative and qualitative evaluations on multiple datasets.
\end{itemize}

\section{Related Work}
\textbf{Deepfake detection.}
Deepfake detection primarily focuses on binary classification, which aims to discriminate the authenticity of a given image or video \cite{pmlricml25yanzhiyuan}.
Various effective strategies have been proposed, including data augmentation \cite{yin2024fine}, biometric information analysis \cite{Xu_2023_ICCV}, and attention mechanisms \cite{Zhao_2021_CVPR}.
Besides, some methods have also investigated the spatial localization task and temporal localization task to cope with different detection scenarios \cite{10702428,xu2025weakly}.
These methods basically provide a discriminative decision for a given face data, lacking additional forensic information for humans. 
Recently, researchers have explored utilizing LLMs to generate the textual reasoning content about forgery traces, which enables the Deepfake detection to be more explainable \cite{guo2025rethinking, yu2025unlocking}.
The LLM-based methods typically fine-tune models using image-text pairs, where the answers usually consist of only text information about forgery traces.
It should be noted that a conventional Deepfake detector is often employed as an expert network to guide the LLM's reasoning process.
In this paper, MARE introduces additional spatial information to motivate the VLM to generate more reliable reasoning content with text-spatial alignment.
\begin{figure*}[htbp]  
    \centering
    \includegraphics[width=\textwidth]{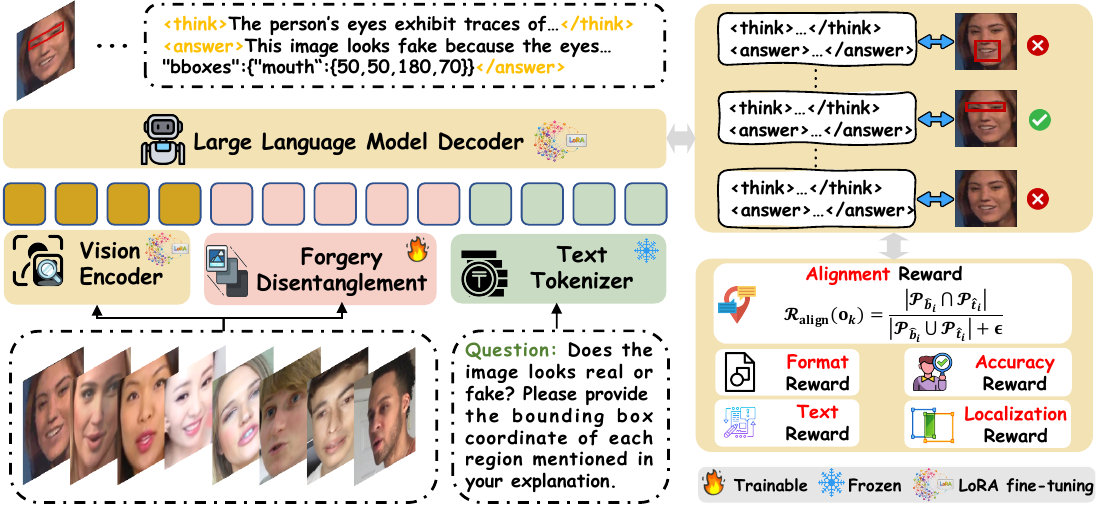}  
    \caption{
        Diagrammatic overview of the proposed MARE framework. 
        The reward functions and RLHF algorithm are displayed on the right side (see Section \ref{sec:reward_func} for details).
        The vision encoder, text tokenizer, and large language model decoder are three key components of VLM.
        The forgery disentanglement module is introduced for intrinsic forgery traces extraction(see Section \ref{sec_FDM} for details).
        During the inference phase, MARE generates text-spatially aligned reasoning content for a given image-text query}
    \label{fig_overview}
\end{figure*}

\textbf{Multimodal alignment in VLMs.}
\label{sec_MA_related}
Alignment is crucial for improving the downstream task accuracy, safety, and reliability of VLMs \cite{li2025survey}.
The general alignment algorithm is reinforcement learning from human feedback (RLHF).
RLHF is a machine learning technique that uses human feedback to train and align models for performing reasoning tasks more aligned with human preferences \cite{pmlr-v235-li24ag, pmlricml25lijinchao}.
The core principle of RLHF is to construct human feedback annotation data and design reward mechanisms.
MM-RLHF \cite{zhang2025mm} constructs a high-quality, fine-grained dataset with 120k preference comparison pairs to advance the alignment of VLMs.
Group relative policy optimization utilizes rule-based reward to improve the models' chain-of-thought ability for more accurate and reliable reasoning content \cite{shao2024deepseekmath}.
RLHF has become one of the most popular and effective paradigms to align VLMs.
However, existing LLM-based Deepfake detectors have insufficient exploration of alignment \cite{meng2025mm}.
In this paper, MARE initially constructs a preference text-spatially aligned dataset.
Based on these data, MARE introduces elaborate reward functions to improve the accuracy and reliability of VLMs in Deepfake detection and reasoning.
\section{Method}
\label{sec_method}
In this section, the specific framework and training details of the proposed MARE are elaborated.
For clarity, an overview of MARE is first introduced.

\subsection{Overview}
\label{sec_overview}
To enhance the accuracy and reliability of VLMs in Deepfake detection and reasoning, we propose MARE under the RLHF paradigm.
The diagrammatic overview is shown in Figure \ref{fig_overview}.
We design comprehensive reward functions and construct a human-preference dataset to incentivize the VLMs to generate text-spatially aligned reasoning content 
\cite{xu2025avatarshield}\footnote{The visual design of this figure is based on AvatarShield \cite{xu2025avatarshield}.}.
Meanwhile, MARE introduces a forgery disentanglement module (FDM) to capture intrinsic forgery traces for precise authenticity identification.

Specifically, we construct a Deepfake multimodal alignment dataset $\mathcal{D}_{ma}=\{\langle v_i, q_i, t_i, b_i \rangle\}_{i=1}^{N}$, where $v_i$, $q_i$, $t_i$, $b_i$ denote the face image, question, textual description, and spatial bounding boxes, respectively.
MARE introduces FDM to disentangle the facial image into identity, structural, and forgery traces features.
The FDM yields a robust authenticity prediction by capturing intrinsic forgery traces.
The prediction serves as an expert prior, which is explicitly embedded into the $q_i$ to guide the VLM's reasoning process.
During the RLHF fine-tuning process, given an input pair $\langle v_i, q_i \rangle$, a set of candidate reasoning responses $\mathcal{O}=\{o_k=(\hat{t}_k, \hat{b}_k)\}_{k=1}^{K}$ will be generated by the VLM.
Leveraging the comprehensive reward functions and the ground truth $\langle t_i, b_i\rangle$, MARE calculates rewards to estimate the relative advantage of each response.
This process drives gradient updates to optimize the policy of VLM, refining its chain-of-thought capability for Deepfake detection and reasoning.

In the inference phase, MARE generates text-spatially aligned reasoning content $\langle \hat{t}_i,\hat{b}_i \rangle$ for a given query $\langle v_i, q_i \rangle$, where $\hat{t}_i$ is the textual reasoning content that contains the authenticity detection result and related explanation, $\hat{b}_i$ provides the spatial bounding boxes of regions mentioned in $\hat{t}_i$.
$\hat{t}_i$ and $\hat{b}_i$ provide mutual validation to improve the accuracy and reliability of reasoning content.

\subsection{Multimodal Alignment}
Alignment could improve the accuracy, safety, and reliability of VLMs in the downstream task \cite{li2025survey}.
In the Deepfake forensic domain, it is particularly crucial to enhance the accuracy and reliability of reasoning content through multimodal alignment.
In this paper, the objective of multimodal alignment is to incentivize the VLM to generate reasoning content that aligns textual and spatial information.
For instance, if the text describes forgery traces about mouth, the reasoning output should provide corresponding bounding boxes for the mouth region.
This mechanism could not only mitigate the VLM's hallucinations but also enhance the reliability of the reasoning content. 

\begin{figure}[t]
  \centering
  \includegraphics[width=\linewidth]{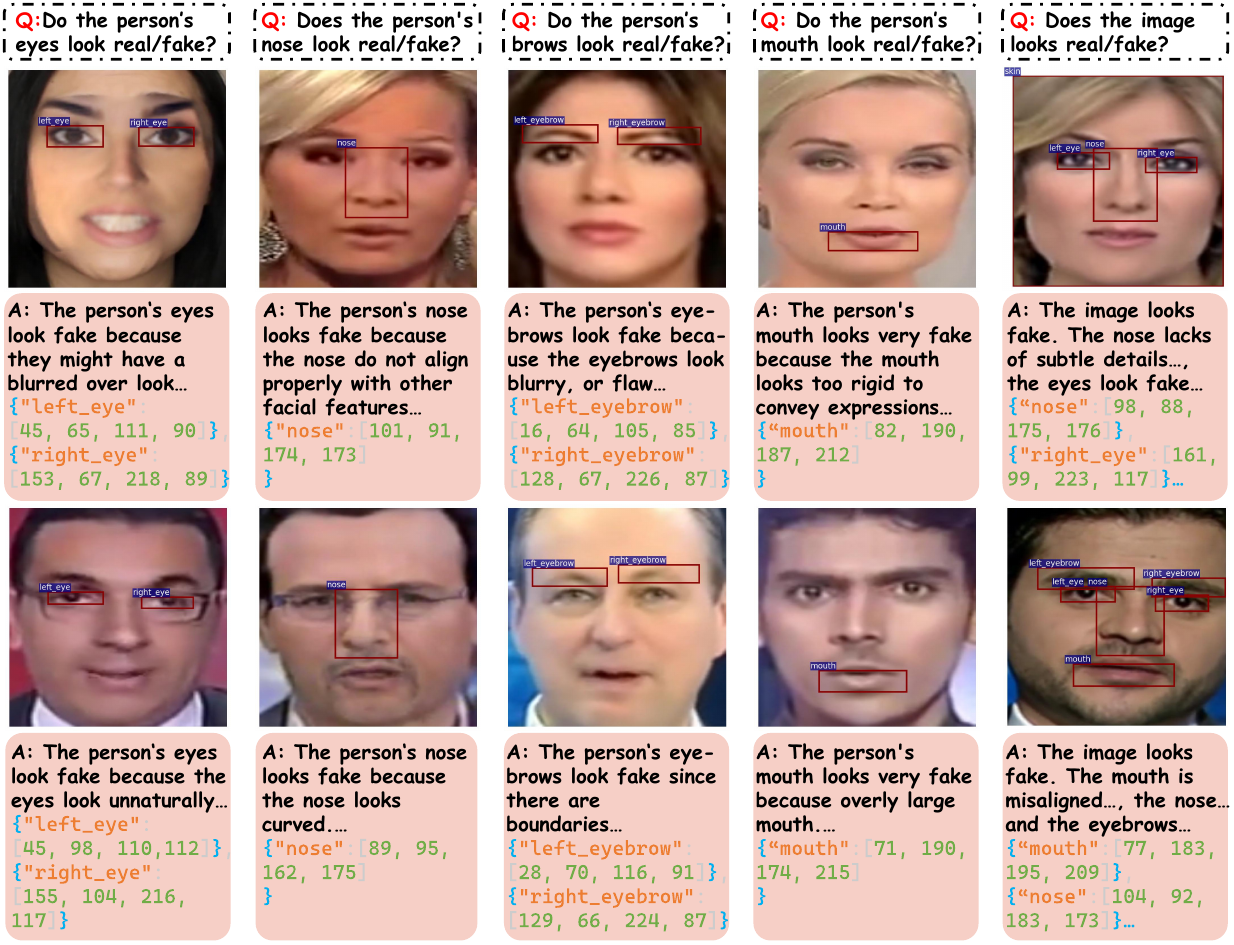}
  \caption{
      Examples of the DMA dataset.
        }
  \label{fig_dataset_DMA}
\end{figure}

Specifically, we construct the Deepfake multimodal alignment dataset (DMA) by augmenting the existing image-text dataset \cite{zhang2024common, yu2025unlocking}, denoted as $\mathcal{D}_{s r c}=\left\{\langle v_i, q_i, t_i\rangle\right\}_{i=1}^N$, where $v_i$ indicates the face image, $q_i$ is a query question, and $t_i$ is the ground-truth textual description about authenticity and detailed analysis.
We propose a two-stage extraction and localization pipeline.
Specifically, a set of facial regions $\mathcal{P}_{key}$ is defined (e.g., eyes, nose, or mouth, see the appendix for details).
We perform keyword retrieval $\mathbb{E}(\cdot,\cdot)$ for each $t_i$ to extract the specific regions $r_i$.
Then, a face landmark detection model $\mathbb{L}(\cdot,\cdot)$ is utilized to locate the spatial bounding boxes $b_i$ from the face image $v_i$.
The pipeline is formulated as:
\begin{equation}
    r_i = \mathbb{E}(t_i, \mathcal{P}_{key}) \quad \quad
    b_i = \mathbb{L}(v_i, r_i)
\end{equation}
Then, the DMA dataset $\mathcal{D}_{ma}=\left\{\langle v_i, q_i, t_i, b_i\rangle\right\}_{i=1}^N$ is obtained, which serves as the foundation for the subsequent multimodal alignment fine-tuning.
The examples of DMA are shown in Figure \ref{fig_dataset_DMA}.

\subsection{Reinforcement Learning Reward Design}
\label{sec:reward_func}
We fine-tune the pre-trained VLMs with the reinforcement learning algorithm (e.g., GRPO \cite{shao2024deepseekmath}).
To incentivize the reasoning capability of VLMs in Deepfake detection and reasoning, we design a set of task-specific reward functions, including format reward $\mathcal{R}_f$, accuracy reward $\mathcal{R}_a$, text relevance reward $\mathcal{R}_t$, ROI reward $\mathcal{R}_r$, and alignment reward $\mathcal{R}_{align}$.

\textbf{Format reward.}
Format reward is to check whether the VLM's responses follow the human preferences format.
Specifically, MARE requires the model to present the reasoning paths enclosed within \texttt{<think>...</think>} tags.
The final reasoning result within \path|<answer>...</answer>| tags should contain two parts: an \texttt{"explanation"} field with a string value describing why the face image is real or fake, and a \texttt{"bboxes"} field listing the bounding box of each region mentioned in the \texttt{"explanation"} that serves as evidence.
Formally,
\begin{equation}
    \mathcal{R}_f(o_k)= \begin{cases}1.0, & \text { if $o_k$ fulfills the format} \\ 0, & \text { otherwise }\end{cases}
\end{equation}
where $o_k$ denotes the $k$-th candidate response for a given fine-tuning instruction $\mathcal{D}_{ma}^{(i)}$.

\textbf{Accuracy reward.}
To support the Deepfake detection and reasoning task, MARE introduces an accuracy reward $\mathcal{R}_a$ based on the correctness of authenticity prediction within the response.
Specifically, we extract keyword \texttt{"fake"} or \texttt{"real"} from the \texttt{"explanation"} field.
Serving as a basic and essential signal, $\mathcal{R}_a$ is to incentivize VLMs explore a valid chain-of-thought path for accurate Deepfake detection.
Formally, 
\begin{equation}
    \mathcal{R}_a(o_k)= \begin{cases}1.0, & \text { if $d_k$ = $g^{(i)}$} \\ 0, & \text { otherwise }\end{cases}
\end{equation}
where $d_k$ denotes the prediction result of authenticity in $o_k$, and $g^{(i)}$ is the corresponding ground truth label in $\mathcal{D}_{ma}^{(i)}$.

\textbf{Text relevance reward.}
Text relevance reward $\mathcal{R}_t$ is designed to incentivize the VLM to generate responses similar to the annotated textual description.
Specifically, we utilize a sentence encoder (e.g., SentenceTransformer \cite{NEURIPS2020_3f5ee243}) to map $t_i$ and $\hat{t}_i^{(k)}$ into embeddings, where $t_i$ is the human annotated textual description in $\mathcal{D}_{ma}^{(i)}$, and $\hat{t}_i^{(k)}$ is the generated textual content in the $k$-th candidate response $o_k$.
$\mathcal{R}_t$ is then obtained by calculating the similarity between these embeddings.
Formally,
\begin{equation}
    \mathcal{R}_t\left(o_k\right)=\max \left(0, \cos \left(\Phi(\hat{t}_i^{(k)}), \Phi(t_{i})\right)\right)
\end{equation}
where $\Phi(\cdot)$ indicates the sentence encoder, $cos(\cdot,\cdot)$ computes the cosine similarity between the two embedding vectors.

\begin{figure}[t]  
    \centering
\includegraphics[width=\linewidth]{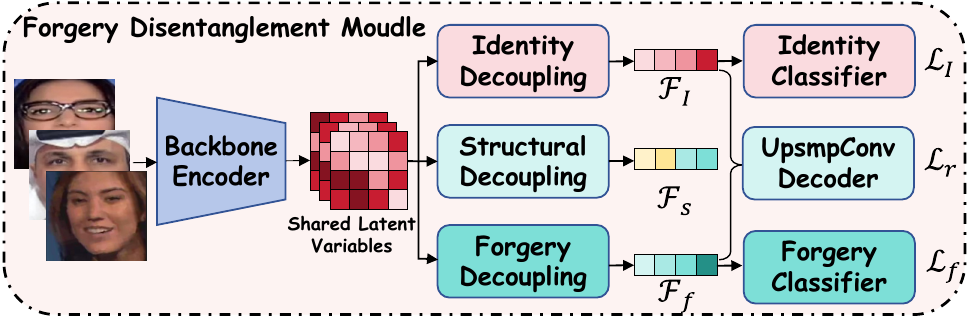}  
    \caption{Illustration of the proposed forgery disentanglement detector (FDM).
The diagrammatic structure of FDM is shown in the left panel, while the right panel depicts the structure of the adversarial classifier.}
    \label{fig_CDD}
\end{figure}

\textbf{Region of interest reward.}
To enhance the VLM's capability in localizing facial regions, MARE introduces a region of interest (ROI) reward $\mathcal{R}_r$.
Specifically, the response $o_k$ is required to contain a \texttt{"bboxes"} field $\hat{b}^{(k)}$, which consists of pairs of facial regions and their corresponding coordinates.
The $\mathcal{R}_r$ evaluates the localization performance to incentivize the VLM to identify the bounding boxes of facial regions.
Formally,
\begin{equation}
    \mathcal{R}_{r}\left(o_k\right)=\frac{1}{|\mathcal{P}|} \sum_{p \in \mathcal{P}} \operatorname{IoU}\left(\hat{b}_{i,p}^{(k)}, b_{i,p}\right)
\end{equation}
where $b_i$ is the spatial bounding boxes in $\mathcal{D}_{ma}^{(i)}$, $\mathcal{P}$ denotes the set of facial regions (e.g., eyes, nose) present in both $\hat{b}_i^{(k)}$ and $b_i$, $\hat{b}_{i,p}^{(k)}$ and $b_{i,p}$ denote the predicted and ground truth bounding boxes for the specific part $p$, respectively.

\textbf{Alignment reward.}
The alignment reward $\mathcal{R}_{align}$ is designed to incentivize the VLM to generate text-spatially aligned reasoning content.
Then MARE could generate more reliable Deepfake detection and reasoning results.
Specifically, for a candidate response $o_k$ of $\mathcal{D}_{ma}^{(i)}$, we extract the set of facial regions $\mathcal{P}_{\hat{b}_i^{(k)}}$ from \texttt{"bboxes"} field, and the set of mentioned facial regions $\mathcal{P}_{\hat{t}_i^{(k)}}$ from the \texttt{"explanation"} field.
The higher the overlap between these two sets, the greater the alignment between textual and spatial reasoning content.
Formally, 
\begin{equation}
    \mathcal{R}_{align}\left(o_k\right)=\frac{\left|\mathcal{P}_{\hat{b}_i^{(k)}} \cap \mathcal{P}_{\hat{t}_i^{(k)}}\right|}{\left|\mathcal{P}_{\hat{b}_i^{(k)}} \cup \mathcal{P}_{\hat{t}_i^{(k)}}\right|+\epsilon}
\end{equation}
where $\left|\cdot\right|$ denotes the cardinality of the set, $\epsilon$ is a constant to prevent division by zero.

\subsection{Forgery Disentanglement Module}
\label{sec_FDM}
Forgery traces in Deepfake face images usually emerge as subtle signals, such as blending boundaries, frequency anomalies, or inconsistent resolution.
Therefore, it is necessary to make the model concentrate on subtle forgery traces rather than strong signals such as identity representation or high-level semantics \cite{dong2023implicit}.
The forgery disentanglement module (FDM) decouples features from the given face image based on representation learning.
The diagrammatic structure of FDM is illustrated in Figure \ref{fig_CDD}.

Specifically, FDM first utilizes a backbone to obtain the shared latent representations $\mathcal{F}^{(i)}$ of the given face image $v_i$.
$\mathcal{F}^{(i)}$ are disentangled into identity features $\mathcal{F}_I$, structural features $\mathcal{F}_s$, and forgery trace features $\mathcal{F}_f$.
To facilitate feature disentanglement, the FDM incorporates dedicated classifiers and loss functions to guide FDM training. 
The identity classifier utilizes $\mathcal{F}_I$ to generate identity predictions, then calculates the identity prediction loss $\mathcal{L}_{I}$ based on the ground truth identity annotations.
Focal loss is utilized to guide the model to focus on challenging samples. Formally,
\begin{equation}
\begin{split}
    \mathcal{L}_{I} &= -\frac{1}{N} \sum_{i=1}^N \sum_{j=1}^{M} \Big[y_{ij}\alpha_j\left(1-\hat{y}_{ij}\right)^{\gamma} \log \left(\hat{y}_{ij}\right)\Big]
\end{split}
\end{equation}
where $N$ indicates the number of samples, $M$ is the number of identities, and $y_{ij} \in \{0,1\}$ is the ground truth identity annotation.
$\hat{y}_{ij}$ denotes the predicted probability that sample $\mathcal{D}_{ma}^{(i)}$ belongs to  the $j$-th identity.
$\alpha_j$ is the balancing parameter for the $j$-th identity, and $\gamma$ is the focusing parameter.

The primary objective of FDM is to capture intrinsic forgery traces from the detected face image.
To provide explicit supervision for this disentanglement, FDM introduces a forgery classifier to perform binary authenticity discrimination based on $\mathcal{F}_f$.
Then, the forgery prediction loss $\mathcal{L}_f$ is obtained.
Formally,
\begin{equation}
\begin{split}
    \mathcal{L}_{f} &= -\frac{1}{N} \sum_{i=1}^N\Big[g_i \cdot \alpha\left(1-\hat{g}_i\right)^{\hat{\gamma}} \log \left(\hat{g}_i\right)+ \\ 
    &\quad \left(1-g_i\right) \cdot(1-\alpha) \hat{g}_i^{\hat{\gamma}} \log \left(1-\hat{g}_i\right)\Big]
\end{split}
\end{equation}
where $g_i \in \{0,1\}$ is the ground truth of authenticity, 
$\hat{g}_i$ denotes the authenticity prediction probability, $\alpha$ and $\hat{\gamma}$ are weighting factors.

Besides identity information and forgery traces, the face image also contains structural information, such as illumination, expressions, and head posture information.
To prevent these factors from interfering with forgery detection, they are disentangled into structural features $\mathcal{F}_s$.
Furthermore, $\mathcal{F}_s$, $\mathcal{F}_I$, and $\mathcal{F}_f$ are concatenated for initial feature reconstruction.
The reconstruction loss $\mathcal{L}_r$ imposes an information constraint that maintains the stability of the disentanglement framework.
Formally,
\begin{equation}
    \mathcal{L}_r=\frac{1}{N} \sum_{i=1}^N\left(\mathcal{F}^{(i)}-\hat{\mathcal{F}}^{(i)}\right)^2
\end{equation}
where $\mathcal{F}^{(i)}$ denotes the shared latent representations within Figure \ref{fig_CDD}, and $\hat{\mathcal{F}}^{(i)}$ is the reconstruction features.
\subsection{Training and Inference}
\textbf{Training.}
MARE contains two training stages: supervised training for the FDM and RL-based optimization for the VLM.
First, the FDM is trained based on existing Deepfake datasets.
The overall loss function is defined as follows:
\begin{equation}
    \mathcal{L} = \lambda_1\mathcal{L}_I + \lambda_2\mathcal{L}_f + \lambda_3\mathcal{L}_r
\end{equation}
where $\lambda_s$ are the hyperparameters to balance each loss component.
Secondly, MARE fine-tunes the VLM based on the constructed DMA dataset via a reinforcement learning algorithm.
The primary objective is to incentivize the VLM's reasoning capabilities in Deepfake detection and reasoning.
The final reward function $\mathcal{R}$ is defined as follows:
\begin{equation}
    \mathcal{R} = \beta_1\mathcal{R}_f+\beta_2\mathcal{R}_a+\beta_3\mathcal{R}_t+\beta_4\mathcal{R}_r+\beta_5\mathcal{R}_{align}
\end{equation}
where $\beta_s$ are the hyperparameters to balance the reward functions.
\begin{table*}[htbp!]
\centering
\caption{The experimental results of intra-dataset detection.
The best results are presented in \textcolor{red}{red}.
The second-best values are in \textcolor{blue}{blue}.
}
\label{intra_dataset_comparison}
\resizebox{0.8\textwidth}{!}{
\begin{tabular}{c|c|cc|cc|cc|cc}
\hline
\multirow{2}{*}{Methods} & \multirow{2}{*}{Venue} & \multicolumn{2}{c|}{FF++} & \multicolumn{2}{c|}{Celeb-DF} & \multicolumn{2}{c|}{WDF} & \multicolumn{2}{c}{DFDC}\\ \cline{3-10} 
 &  & \multicolumn{8}{c}{\textit{Metric: Acc (\% $\uparrow$) / AUC (\% $\uparrow$)}} \\ \hline
MultiAtt \cite{Zhao_2021_CVPR} & CVPR21 & 97.60 & 99.29 &  97.92 & \textcolor{blue}{99.94} & 82.86 & 90.71  & -- &-- \\
 RECCE \cite{Cao_2022_CVPR} & CVPR22 & 97.06 & 99.32 &  98.59 & \textcolor{blue}{99.94} & 83.25 & 92.02  & \textcolor{blue}{81.20} & 91.33\\
TALL \cite{Xu_2023_ICCV} & ICCV23 & \textcolor{blue}{98.65} & \textcolor{red}{99.87} & 97.57 & 98.55 & -- & --  & -- & 76.78\\
TGN \cite{10654318} & TIFS24 & -- & 99.32 & -- & 99.59 & -- & --  & -- & \textcolor{blue}{96.24}\\
M2F2 \cite{guo2025rethinking} & CVPR25 & \textcolor{red}{98.79} & \textcolor{blue}{99.34} &  \textcolor{blue}{98.98} & 99.92 & \textcolor{blue}{86.05} & \textcolor{blue}{93.14}  &  -- & --\\ \hline
MARE &  & 96.55 & 98.86 & \textcolor{red}{99.61} & \textcolor{red}{99.98} & \textcolor{red}{87.22} & \textcolor{red}{93.70}  & \textcolor{red}{93.75} & \textcolor{red}{98.30} \\ \hline
\end{tabular}
}
\end{table*}

\textbf{Inference.}
Given the face image and question prompts, MARE generates a binary authenticity detection result, followed by an explainable textual description with aligned spatial bounding boxes, as shown in Figure \ref{fig_overview}.

\begin{table*}[htbp]
\centering
\caption{The experimental results of fuse-dataset detection.
The best results are presented in \textcolor{red}{red}.
The second-best values are in \textcolor{blue}{blue}.}
\label{fuse_dataset_comparison}
\resizebox{\textwidth}{!}{
\begin{tabular}{c|c|cc|cc|cc|cc|cc}
\hline
Methods & Venue & \multicolumn{2}{c|}{FF++} & \multicolumn{2}{c|}{Celeb-DF} & \multicolumn{2}{c|}{WDF} & \multicolumn{2}{c|}{DFDC} & \multicolumn{2}{c}{DFD}\\ \cline{3-12} 
 &  & \multicolumn{10}{c}{\textit{Metric: Acc (\% $\uparrow$) / AUC (\% $\uparrow$)}} \\ \hline
EfficientNet \cite{tan2019efficientnet} & ICML2019 & 91.19 & 94.18 & 98.65 & 99.87 & 80.02 & 88.22 & 85.94 & 93.84  & 95.63 & 98.14 \\
ExpD \cite{ba2024exposing}& AAAI2024 & 95.60 & 98.40 & 99.42 & \textcolor{blue}{99.99} & 80.89 & 88.02 & \textcolor{blue}{96.09} & \textcolor{blue}{99.30} & \textcolor{blue}{97.53} & \textcolor{blue}{99.23} \\ 
SFIC \cite{10286083}& TIFS2024 & 90.24 & 90.30 & 92.86 & 97.76 & 76.55 & 84.37 & 84.38 & 91.28 & 93.74 & 94.57 \\ 
M2F2 \cite{guo2025rethinking} & CVPR2025 & \textcolor{blue}{97.38} & \textcolor{blue}{99.25} & \textcolor{blue}{99.61} & \textcolor{blue}{99.99} & \textcolor{blue}{84.86} & \textcolor{red}{92.70} & 95.31 & 99.16  &  96.51 & 98.67 \\ \hline
MARE &  & \textcolor{red}{97.50} & \textcolor{red}{99.28} & \textcolor{red}{100.00} & \textcolor{red}{100.00} & \textcolor{red}{87.72} & \textcolor{blue}{92.66}
& \textcolor{red}{98.83} & \textcolor{red}{99.77} & \textcolor{red}{98.25} & \textcolor{red}{99.68} \\ \hline
\end{tabular}%
}
\vspace{-0.5em}
\end{table*}
\begin{table}[!t]
\centering
\caption{The identification performance of generated textual description on the DMA dataset.}
\label{tab_identifaication_performance}
\resizebox{0.8\linewidth}{!}{
\begin{tabular}{c|c|cc}
\hline
\multirow{2}{*}{Methods} & \multirow{2}{*}{Fine-tuned} & \multicolumn{2}{c}{Detection} \\ 
 &  & Acc & F1 \\ \hline 
Qwen & \ding{55} & 59.41 & 51.04 \\
InternVL & \ding{55} & 42.33 & 47.35 \\
LLaVA & \ding{55} & 51.41 & 37.04 \\
LLaVA & \ding{51} & 86.41 & 92.10 \\
DDVQA-BLIP & \ding{51} & 87.49 & 90.07 \\
M2F2 & \ding{51} & 95.23 & 96.61 \\
\hline
\rowcolor{gray!20} MARE & \ding{51} & 98.09 & 97.04
\\ \hline
\end{tabular}%
}
\end{table}

\section{Experiment}
\subsection{Experimental Setup}
\textbf{Datasets.}
We evaluate MARE on both detection and reasoning generation tasks\footnote{}.
The existing Deepfake datasets FaceForensics (FF++) \cite{Rossler_2019_ICCV}, Celeb-DF \cite{Li_2020_CVPR}, WildDeepfake (WDF) \cite{10.1145/3394171.3413769}, DFDC \cite{dolhansky2020deepfake}, and DFD \cite{DFD2019deepfake} are used for detection tasks.
The augmented Deepfake multimodal alignment dataset (DMA), based on DDVQA \cite{zhang2024common}, is utilized for the reasoning generation task.

\textbf{Evaluation metrics.}
In this paper, we utilize Area Under the Curve (AUC) and accuracy (Acc) to measure the binary authenticity detection performance.
To evaluate the performance of the reasoning content of VLMs, we measure the identification Acc and F1 by extracting keywords (\texttt{"fake"} or \texttt{"real"}) from the generated textual description.

\textbf{Implementation details.}
The VLM-R1 \cite{shen2025vlm} is utilized as the toolkit to fine-tune the VLMs.
The reinforcement learning algorithm employs GRPO \cite{shao2024deepseekmath}.
The backbone encoder of FDM is Xception \cite{chollet2017xception}.
The fine-tuned VLMs are Qwen2.5-VL-3B \cite{bai2025qwen2}, Qwen2.5-VL-7B, and InternVL2.5-4B \cite{chen2024expanding}.
The face landmark detection model $\mathbb{L}(\cdot)$ is MediaPipe Face Mesh.
More details of the implementation are presented in the appendix.

\subsection{Detection Performance}
In this section, we compare the proposed MARE with previous state-of-the-art methods under intra- and fuse-dataset setups, respectively.
For intra-dataset comparisons, we cite baseline results from their original papers.
For the fuse-dataset experiments, all methods are retrained for fair comparison.

\textbf{Intra-dataset performance.}
As shown in Table \ref{intra_dataset_comparison}, the experimental results show that MARE achieves relatively good performance on both Acc and AUC under intra-dataset comparison.
On Celeb-DF, WDF, and DFDC datasets, MARE has achieved performance improvements in both Acc and AUC.
It could be observed that MARE achieves state-of-the-art performance on the challenging datasets WDF and DFDC, while it underperforms on the FF++ dataset.
This phenomenon could be attributed to the characteristics of FF++, which is generated under five specific Deepfake methods (Deepfakes, Face2Face, FaceShifter, FaceSwap, and NeuralTextures) and constrained scenarios (interviews or news broadcasting).
The models may exploit implicit scenario- or identity-specific information to improve the detection performance.
In contrast, the FDM of MARE is designed to capture intrinsic forgery traces that are disentangled from image semantics.
The experimental results indicate MARE outperforms on both challenging datasets WDF and DFDC.
This demonstrates that MARE effectively decouples forgery traces from the high-level semantic information, enabling its superior performance in complex Deepfake detection scenarios.

\textbf{Fuse-dataset performance.}
To evaluate the proposed MARE's ability to extract intrinsic forgery traces from different Deepfake scenarios, we introduce the fuse-dataset experiment.
In the fuse-dataset experiment, the training subsets of five Deepfake datasets (including FF++, Celeb-DF, WDF, DFDC, and DFD) are merged to create a new training set, and evaluation is conducted on the test subset of each Deepfake dataset individually.
To ensure a fair comparison, all comparison methods are retrained based on their official open-source implementations.
The fuse-dataset experimental results are shown in Table \ref{fuse_dataset_comparison}.
It could be observed that MARE achieves consistent performance improvements on the FF++, Celeb-DF, WDF, DFDC, and DFD datasets.
In contrast, comparison method M2F2 shows improved detection performance on Celeb-DF, but exhibits performance degradation on the FF++ and WDF.
The primary reason may be the implicit leakage of scenario and identity information, which has impacted its performance in complex Deepfake detection scenarios.
The experimental results indicate that MARE could capture the intrinsic forgery traces when faced with diverse Deepfake scenarios.
Also, this demonstrates that MARE could decouple intrinsic forgery traces from high-level and diverse semantics.

\subsection{Reasoning Performance}
\label{sec_experiment_attri}
In this section, we conduct both quantitative and qualitative evaluations of the reasoning content generated by MARE.
Due to space constraints, some evaluations are presented in the appendix.

\textbf{Quantitative performance.}
To quantitatively evaluate the generated textual description, we extract identification keywords (e.g., \texttt{"fake"} or \texttt{"real"}) from the reasoning content. 
Acc and F1 are calculated based on the ground truth labels.
The identification performance is shown in Table \ref{tab_identifaication_performance}.

The pre-trained Qwen \cite{bai2025qwen2}, InternVL \cite{chen2024expanding}, and LLaVA \cite{NEURIPS2023_6dcf277e} are utilized as baseline models.
The results indicate the pre-trained VLMs underperform on Deepfake detection and reasoning.
The primary reason is that forged face images are often visually difficult to distinguish from genuine ones, since the forgery traces are typically imperceptible.
After supervised fine-tuning (SFT) with annotated data, LLaVA achieves significant improvements in both Acc and F1 scores.
DDVQA-BLIP \cite{zhang2024common} and M2F2 combined with SFT to enhance the reasoning capability of VLMs, which achieves further performance improvements.
However, the alignment problem in the reasoning process has not been sufficiently explored.
MARE constructs a multimodal alignment dataset and designs comprehensive reward functions, and utilizes reinforcement learning to incentivize the VLM to generate accurate and reliable reasoning content.
That is the reason that MARE achieves superior performance on both Acc and F1 scores.
\begin{figure}[!t]
  \centering
  \includegraphics[width=\linewidth]{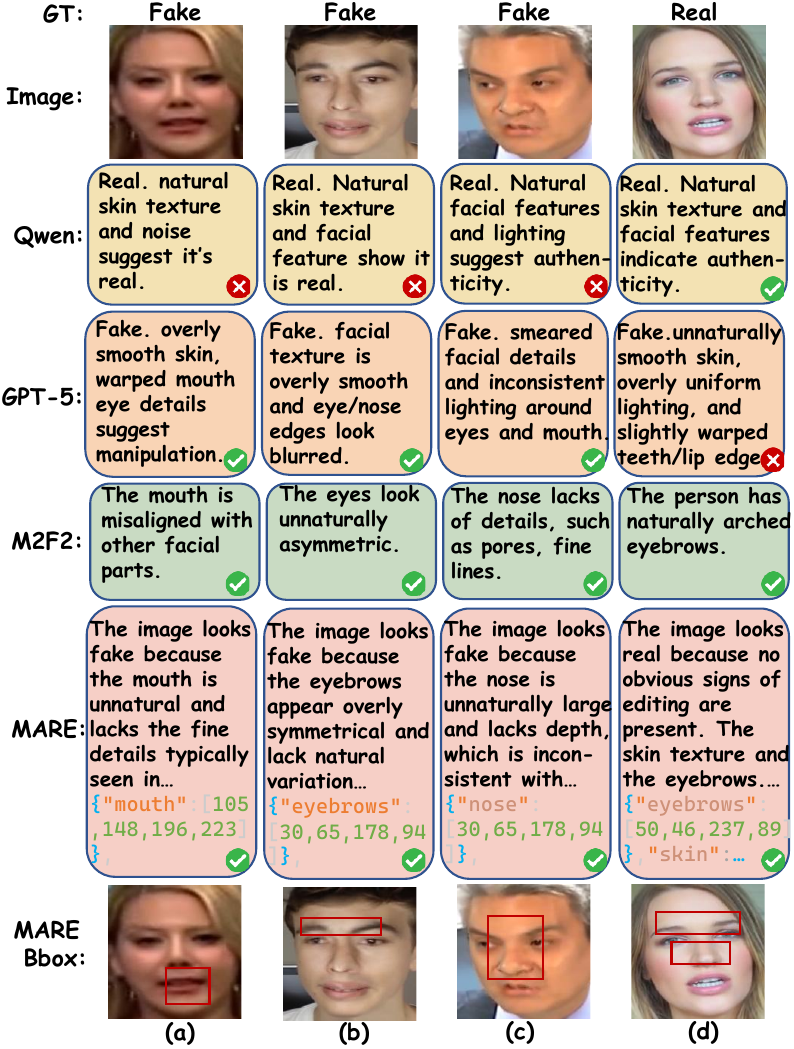}
  \caption{
      Examples of reasoning content generated by Qwen, GPT-5, M2F2, and MARE.
        }
  \label{fig_MARE_samples}
  \vspace{-0.6em}
\end{figure}

\begin{figure}[thbp]
  \centering
  \includegraphics[width=0.8\linewidth]{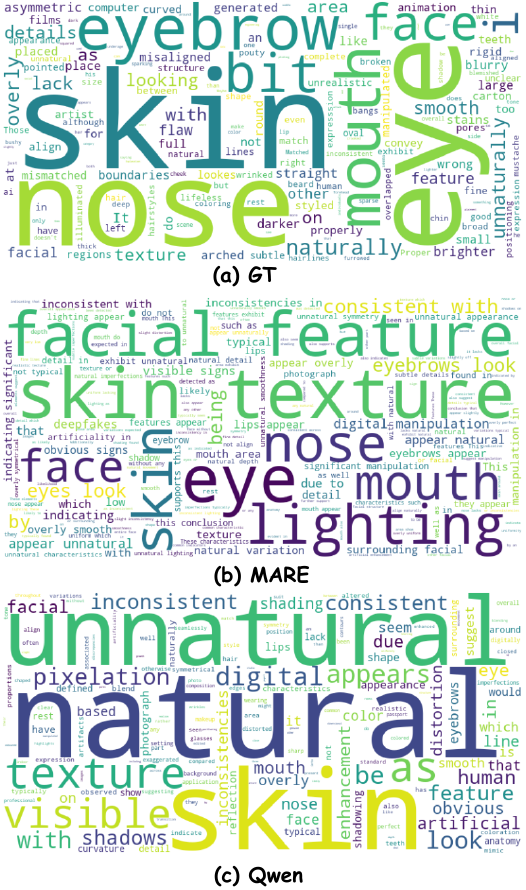}
  \caption{Visualization of word clouds derived from human annotated text (GT), MARE's reasoning content, and Qwen's reasoning content.
        }
  \label{fig_appendix_wordCloud}
\end{figure}

\textbf{Qualitative performance.}
Figure \ref{fig_MARE_samples} exhibits the reasoning content generated by different models.
We employ the API of Qwen-VL and GPT-5 with the prompt \texttt{"Does this 
image look fake/real?"}.
Qwen identifies all samples as \texttt{"real"}, and generates the textual description that has a generalized and templated style.
GPT-5 classifies all samples as \texttt{"fake"}.
Compared to Qwen, the textual description generated by GPT-5 captures more details, such as eyes, nose, and teeth edges.
The results in Figure \ref{fig_MARE_samples} and Table \ref{tab_identifaication_performance} indicate that the existing general-purpose VLMs have unsatisfactory accuracy in face image authenticity identification.
M2F2 achieves improved detection accuracy by introducing expert knowledge.
Besides, M2F2 generates fine-grained textual descriptions based on a detailed prompt (e.g., Figure \ref{fig_MARE_samples} (a) utilizes \texttt{"The mouth in the image is fake, tell me why$...$"} as the prompt).
In contrast, MARE produces reasoning content with text-spatial alignment, which is more aligned with human preferences.
It should be noted that MARE uses the prompt \texttt{"Does this image look fake/real$...$?"}, which does not mention specific facial regions.
Despite this, MARE could still perceive, locate, and analyze the detected face image, thus generating more accurate and reliable reasoning content.

\textbf{Visualization of word clouds}
To provide more intuitive analysis of explainable textual description, Figure \ref{fig_appendix_wordCloud} visualizes the word clouds generated from human annotated text, MARE's reasoning content, and Qwen's reasoning content.
It could be observed that both human annotated text and MARE's reasoning focus on analyzing specific facial regions.
For example, \texttt{"nose", "eyes"}, and \texttt{"mouth"} are the high-frequency words in the word clouds of GT and MARE.
In contrast, the Qwen exhibits limited attention to the facial regions during the reasoning process.
Figure \ref{fig_appendix_wordCloud} indicates the reasoning content generated by MARE for Deepfake detection and reasoning aligns more closely with human preferences.
\subsection{Ablative Study and Analysis}
This section conducts ablation studies to analyze the impact of the VLM backbone within MARE and the specific loss components in the FDM.
First, we compare the performance of different VLM architectures.
Then, based on the best-performing model, we analyze the contribution of $\mathcal{L}_I$ and $\mathcal{L}_r$ within the FDM.

\textbf{Ablation of VLM architecture.}
To evaluate the performance of different VLMs in Deepfake detection and reasoning, we conduct ablation studies on three VLM architectures (InternVL2.5-4B, Qwen2.5-VL-3B, and Qwen2.5-VL-7B) based on DMA dataset.
In this ablation study, all experimental settings except the VLM architecture remain fixed, including the FDM and reward functions. 
The experimental results are shown in Table \ref{tab_ablation_vlms}.
Qwen2.5-VL models show better performance than the InternVL2.5.
This is mainly because Qwen2.5-VL models hold advantages in complex tasks due to their stronger cross-modal understanding abilities.
Besides, Qwen2.5-VL-7B has a superior capability than Qwen2.5-VL-3B under the influence of the scaling law.

\textbf{Ablation of loss components in FDM.}
To verify the effectiveness of the specific loss components within the FDM, we conduct ablation studies on the identity loss $\mathcal{L}_I$ and reconstruction loss $\mathcal{L}_r$.
These experiments are performed based on the best-performing Qwen2.5-VL-7B model, and evaluated on the two challenging WDF and DFDC datasets under the fuse-dataset setup.
The results are shown in Table \ref{tab_ablation_fdm}.
It could be observed that the performance improves on the WDF dataset after introducing the $\mathcal{L}_I$.
This indicates that explicit identity supervision helps MARE to decouple identity information from forgery traces features, reducing the interference of identity signals on authenticity identification.
Besides, MARE achieves the best performance on both WDF and DFDC after the introduction of $\mathcal{L}_r$, which demonstrates that $\mathcal{L}_r$ imposes a crucial information constraint that ensures the stability of the disentanglement architecture.
Therefore, both $\mathcal{L}_I$ and $\mathcal{L}_r$ help to improve the ability of MARE in capturing intrinsic forgery traces features. 
\begin{table}[!t]
\centering
\caption{The identification performance of different VLM architectures base on DMA dataset.}
\label{tab_ablation_vlms}
\resizebox{0.65\linewidth}{!}{
\begin{tabular}{c|cc}
\hline
\multirow{2}{*}{VLM}  & \multicolumn{2}{c}{Detection} \\ 
   &   Acc & F1 \\ \hline 
  InternVL2.5-4B & 94.27 & 90.68 \\
  Qwen2.5-VL-3B& 95.42 & 92.68 \\
\rowcolor{gray!20} Qwen2.5-VL-7B & 98.09 & 97.04
\\ \hline
\end{tabular}%
}
\end{table}
\begin{table}[!t]
\centering
\caption{Ablation study on the specific loss components of the FDM.}
\label{tab_ablation_fdm}
\resizebox{0.8\linewidth}{!}{
\begin{tabular}{cc|cccc}
\hline
\multirow{2}{*}{$\mathcal{L}_I$} & \multirow{2}{*}{$\mathcal{L}_r$} & \multicolumn{2}{c}{WDF} &  \multicolumn{2}{c}{DFDC}\\ 
 &    & Acc & F1 & Acc & F1\\ \hline 
 &  & 83.62 & 84.62 & 94.92 & 96.07\\
\ding{51}&  & 85.48 & 86.15 & 93.75 & 95.24\\
\rowcolor{gray!20}  \ding{51}& \ding{51}&  87.72 & 87.82 & 98.83 & 99.08 \\
 \hline
\end{tabular}%
}
\vspace{-1em}
\end{table}

\begin{figure}[htbp]
  \centering
  \includegraphics[width=\linewidth]{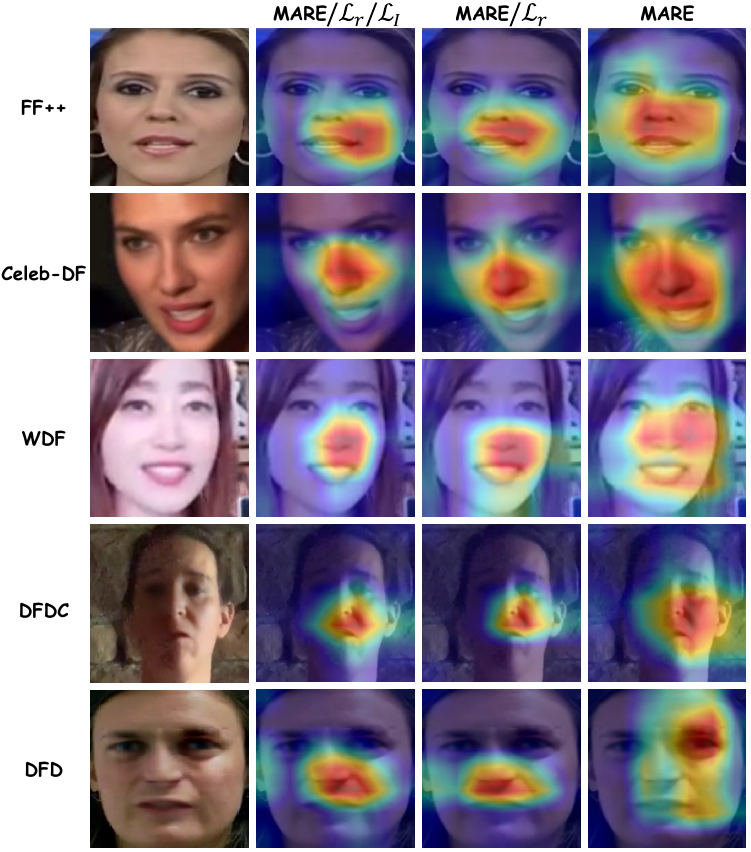}
  \caption{
      Examples of forged attention maps from FF++, Celeb-DF, WDF, DFDC, and DFD dataset.
      Column MARE/$\mathcal{L}_r$ indicates the attention maps of MARE removing loss $\mathcal{L}_r$.
      Column MARE/$\mathcal{L}_r$/$\mathcal{L}_I$ indicates the attention maps of MARE removing loss $\mathcal{L}_r$ and $\mathcal{L}_I$.
        }
  \label{fig_appendix_vis}
\end{figure}

\textbf{Visualization of forged attention maps.}
To more vividly display MARE's attention maps when capturing forgery traces, we visualize some examples in this section.
As shown in Figure \ref{fig_appendix_vis}, it could be observed that MARE focuses not only on specific facial regions but also on facial boundaries.
During the Deepfake synthesis process, the blending of source and target faces typically leaves intrinsic forgery traces at the facial boundaries.
The attention maps in Figure \ref{fig_appendix_vis} indicate that MARE attempts to capture intrinsic forgery traces from multiple perspectives to enhance the accuracy of Deepfake authenticity identification.

\section{Conclusion}
In this paper, we propose MARE, a novel framework designed to enhance the accuracy and reliability of VLMs in Deepfake detection and reasoning.
MARE aims to generate text-spatially aligned reasoning content within the RLHF paradigm.
We first construct a human feedback annotation dataset and design the reward mechanism.
A Deepfake multimodal alignment dataset is introduced to augment the existing image-text Deepfake dataset with aligned spatial localization information.
To effectively incentivize the reasoning capability of VLM, MARE designs a comprehensive reward mechanism comprising five functions: format, accuracy, text relevance, ROI, and alignment rewards.
Besides, a forgery disentanglement module is introduced to capture intrinsic forgery traces for precise authenticity identification, which disentangles the face image into identity, structural, and forgery traces features.
Extensive quantitative and qualitative experiments validate that MARE achieves state-of-the-art performance in Deepfake detection and reasoning.
We observe that existing evaluation methodologies, such as accuracy, AUC, and NLP metrics, are becoming insufficient for assessing the logical consistency and explainability of explainable Deepfake detection. In the future, we will explore more effective evaluation methodologies for generative Deepfake reasoning content.

\begin{acks}
This work is supported by the National Natural Science Foundation of China (No.62441237).
\end{acks}
\bibliographystyle{ACM-Reference-Format}
\bibliography{sample-base}


\end{document}